\documentclass[10pt,twocolumn,letterpaper]{article}

\usepackage[pagenumbers]{cvpr} 

\definecolor{cvprblue}{rgb}{0.21,0.49,0.74}
\usepackage[pagebackref,breaklinks,colorlinks,allcolors=cvprblue]{hyperref}
\usepackage{booktabs}
\usepackage{amssymb}
\usepackage{multirow}
\definecolor{LightBlue}{rgb}{0.9,0.94,1}
\usepackage{wrapfig}

\def\paperID{5930} 
\def\confName{CVPR}
\def\confYear{2025}
\title{ComposeAnyone: Controllable Layout-to-Human Generation with Decoupled Multimodal Conditions}

\author{
    \textbf{Shiyue Zhang}\textsuperscript{\rm 1*}, 
    \textbf{Zheng Chong}\textsuperscript{\rm 1,4*},
    \textbf{Xi Lu}\textsuperscript{\rm 1}, 
    \textbf{Wenqing Zhang}\textsuperscript{\rm 2}, 
    \textbf{Haoxiang Li} \textsuperscript{\rm 3},\\
    \textbf{Xujie Zhang}\textsuperscript{\rm 1}, 
    \textbf{Jiehui Huang}\textsuperscript{\rm 1}, 
    \textbf{Xiao Dong}\textsuperscript{\rm 1},
    \textbf{Xiaodan Liang}\textsuperscript{\rm 1,4$\dagger$}
    \vspace{2mm}
    \\
    Sun Yat-Sen University\textsuperscript{\rm 1}, National University of Singapore\textsuperscript{\rm 2}, \\
    Pixocial Technology\textsuperscript{\rm 3}, Pengcheng Laboratory\textsuperscript{\rm 4}   
    \vspace{1mm}\\
    \small\texttt{* Equal Contribution,    $\dagger$ Corresponding Author} \vspace{1mm}\\
    \noindent\makebox[\linewidth][c]{\normalsize\href{https://github.com/Zhangshy1019/ComposeAnyone}{\texttt{https://github.com/Zhangshy1019/ComposeAnyone}}}
}

\begin{document}
\twocolumn[{
    \renewcommand\twocolumn[1][]{#1}%
    \maketitle
    \vspace{-5mm}
    \centering
    \captionsetup{type=figure}
    \includegraphics[width=\textwidth]{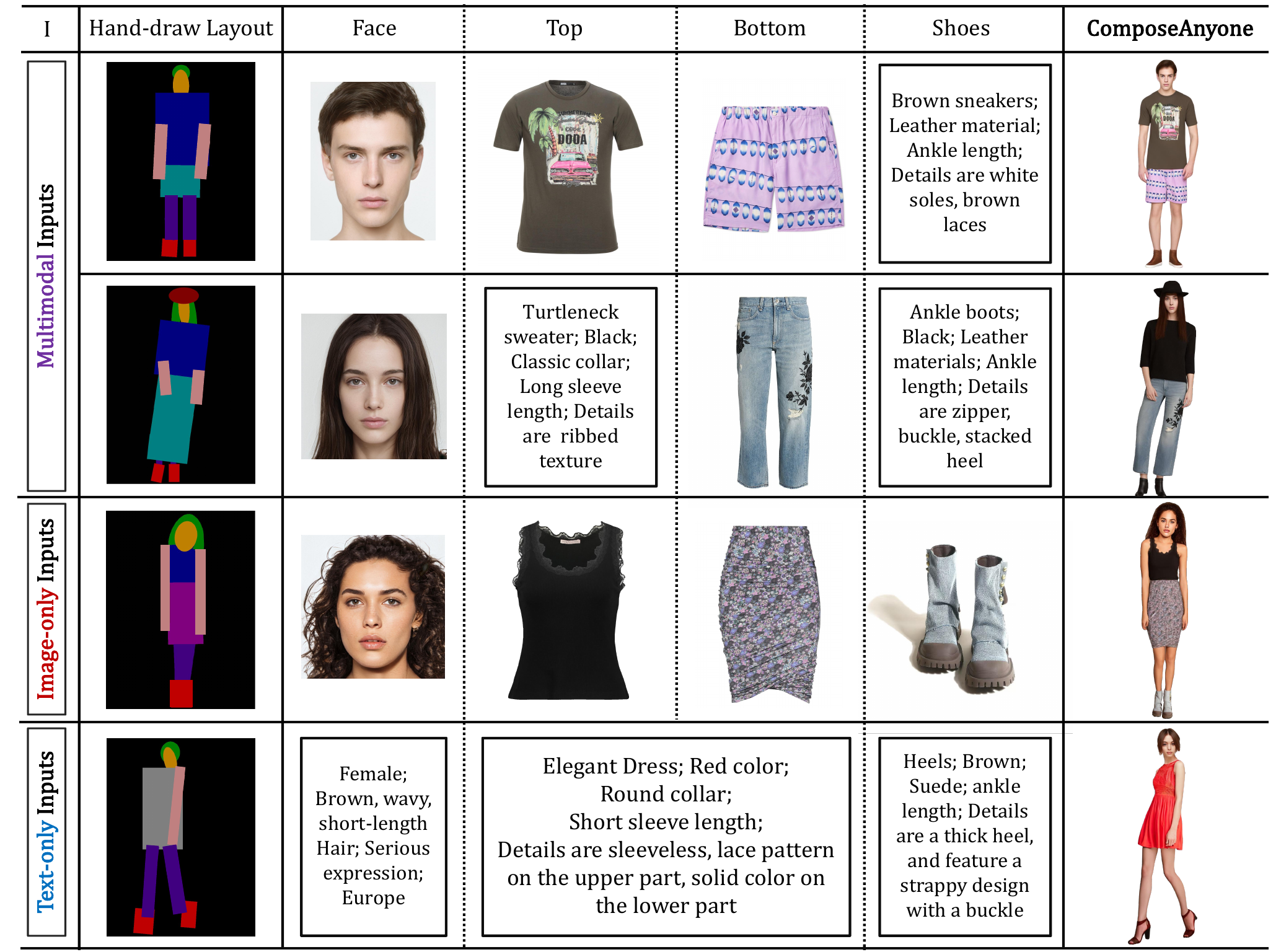}
    \captionof{figure}{ ComposeAnyone is capable of generating high-quality human images with decoupled multimodal conditions, such as captions, reference images, and hand-drawn layouts.
    }
    \label{fig:teaser}
    \vspace{4mm}
}]
\begin{abstract}
Building on the success of diffusion models, significant advancements have been made in multimodal image generation tasks. Among these, human image generation has emerged as a promising technique, offering the potential to revolutionize the fashion design process. 
However, existing methods often focus solely on text-to-image or image reference-based human generation, which fails to satisfy the increasingly sophisticated demands. To address the limitations of flexibility and precision in human generation, we introduce ComposeAnyone, a controllable layout-to-human generation method with decoupled multimodal conditions. 
Specifically, our method allows decoupled control of any part in hand-drawn human layouts using text or reference images, seamlessly integrating them during the generation process. The hand-drawn layout, which utilizes color-blocked geometric shapes such as ellipses and rectangles, can be easily drawn, offering a more flexible and accessible way to define spatial layouts. Additionally, we introduce the ComposeHuman dataset, which provides decoupled text and reference image annotations for different components of each human image, enabling broader applications in human image generation tasks. 
Extensive experiments on multiple datasets demonstrate that ComposeAnyone generates human images with better alignment to given layouts, text descriptions, and reference images, showcasing its multi-task capability and controllability. 

\end{abstract}    
\section{Introduction}
\label{sec:intro}
Recent advances in diffusion models have catalyzed significant breakthroughs in generative AI, especially revealed unprecedented potential in vision tasks incorporating various controls during image generation, such as layout-guided text-to-image generation~\cite{zheng2023layoutdiffusion,li2023gligen,yang2023reco,avrahami2023spatext,densediffusion,bar2023multidiffusion,chen2024training, wang2024instancediffusion,zhou2024migc} and subject-driven image customization~\cite{chen2023photoverse,xiao2024fastcomposer,wang2024instantid,gal2022image,kumari2023multi,hu2021lora,ruiz2023dreambooth,shi2024instantbooth,ye2023ip-adapter,patel2024lambda,wei2023elite,yuan2023customnet,chen2023anydoor}. Building on these strides, human generation~\cite{sarkar2021humangan,jiang2022text2human,zhang2022humandiffusion,cheong2022pose,fu2022stylegan} has gained traction as a compelling research area with broad applications in fashion design, virtual character creation, and social media content. 

Nevertheless, while most existing methods~\cite{rombach2021highresolution,sarkar2021humangan,chen2023anydoor,cheong2022pose} perform well with single-modality inputs, they have not been extended to handle multimodal inputs. As a result, they exhibit clear limitations in complex scenarios that involve multiple sources of information, especially in tasks requiring advanced multimodal collaboration. For example, most subject-driven methods~\cite{ye2023ip-adapter,yuan2023customnet,wei2023elite,chen2023anydoor} are designed to accommodate single-image inputs, with alternatives like $\lambda$-ECLIPSE~\cite{patel2024lambda} supporting sequential multi-image inputs, which remain cumbersome for users. Additionally, methods~\cite{li2023gligen,ye2023ip-adapter,patel2024lambda,yuan2023customnet,wei2023elite} often rely on paired text-image inputs, increasing complexity and redundancy, and potentially leading to neglect of conditions. This makes it challenging to achieve both high fidelity and diversity in multimodal human figure generation. Therefore, there is an urgent need for technological breakthroughs to enable more efficient and accurate data fusion in such scenarios. More flexible, intuitive, and user-friendly human generation methods capable of seamlessly integrating multimodal conditions should be explored.

To address these challenges, we introduce ComposeAnyone, a novel multimodal human generation method designed to enhance flexibility and precision under decoupled multimodal input conditions. ComposeAnyone introduces the concept of "hand-draw layouts," allowing users to specify spatial positions of individual human components through color-block drawings composed of simple geometric forms, such as ellipses and rectangles. This approach leverages distinct color information to delineate regions, effectively preventing feature confusion among different human components and achieving more accurate spatial layout control. 
Our key innovation lies in flexible multimodal input, enabling the selection of either text or images for human component descriptions, facilitating non-paired inputs, as illustrated in \Cref{fig:teaser}. Moreover, we support the integration of multiple reference images for pixel-level fusion, thereby obviating the need for sequential input processing. 
Building upon this foundation, we designed a data-decoupled pipeline that integrates text, reference images, and hand-draw layouts. By spatially aligning latent features, this model produces human images that more closely align with the intended multimodal conditions in both content and structure. Additionally, we apply attention modulation during inference to further enhance spatial coherence and textual consistency.
Besides, we construct a multimodal dataset—ComposeHuman, encompassing human images, hand-drawn layouts, fine-grained textual descriptions, and human component assemblies. 
Experimental results reveal that ComposeAnyone outperforms existing methods in both flexibility and quality, particularly excelling in diverse human figure generation tasks. Its high customizability also allows users to add or remove accessories (e.g., hats, bags) by simply adjusting the hand-draw layout, enabling real-time, personalized image creation. 

Our contributions are summarized as follows:
\begin{itemize}

\item  We propose ComposeAnyone, a novel controllable Layout-to-Human generation method that utilizes hand-drawn geometric shapes as layout, combined with textual and image references for different human components, to generate highly consistent and realistic human images for text-only, image-only, or mix-modal tasks. 

\item  We construct a multimodal decoupled Layout-to-Human dataset, called ComposeHuman, by annotating each body part with text labels and semi-supervised reference image extraction to obtain decoupled multimodal references. Additionally, we convert traditional segmentation into hand-drawn layouts, allowing for more flexible and detailed spatial organization. 

\item Extensive experiments on layout-guided generation and subject-driven generation tasks, conducted on VITON-HD, DressCode, and DeepFashion datasets, demonstrate that ComposeAnyone can generate human images that better align with the given layout, text descriptions, and reference images, highlighting its multi-task capability and controllability. 
    
    
\end{itemize}

\section{Related Work}
\label{sec:related_work}
\subsection{Layout-Guided Image Generation}
Layout-to-Image (L2I) generation seeks to synthesize images from layouts, using category-labeled bounding boxes as structural guidance—essentially reversing the object detection process. 
Earlier methods~\cite{sun2019layout1,li2021layout2,he2021layout3,wang2022layout4,sylvain2021layout5} primarily utilized generative adversarial networks (GANs). Amid the surge of diffusion-based generative methods, integrating layout into diffusion process~\cite{cheng2023layoutdiffuse,zheng2023layoutdiffusion,li2023gligen,xie2023boxdiff,xiao2023r&b,yang2023reco,avrahami2023spatext,densediffusion,bar2023multidiffusion,chen2024training, wang2024instancediffusion,zhou2024migc} has markedly expanded controllability of generated images. 
LayoutDiffusion~\cite{zheng2023layoutdiffusion} integrates a patch-based fusion strategy, GLIGEN~\cite{li2023gligen} incorporates grounded embeddings within gated Transformer layers, MIGC~\cite{zhou2024migc} applies an instance-enhanced attention mechanism for high-precision shading, and InstanceDiffusion~\cite{wang2024instancediffusion} enables diverse forms of spatial control.
However, these methods are limited to text input, resulting in a rather monotonous task and generation process. Our proposed approach enables flexible, customized design by integrating textual description, reference image, and layout inputs across multiple modalities to get a more fine-grained image.

\subsection{Subject-Driven Image Customization}
Subject-driven image generation focuses on the fluid integration of target subject attributes into novel scenes or perspectives~\cite{chen2023photoverse,xiao2024fastcomposer,wang2024instantid,gal2022image,kumari2023multi}, ensuring coherence and consistency.
LoRa~\cite{hu2021lora} and DreamBooth~\cite{ruiz2023dreambooth} finetune pre-trained models for each object individually at test-time, enabling subject-specific generation but tend to be overfitted and time-consuming. 
IP-Adapter~\cite{ye2023ip-adapter}, ELITE~\cite{wei2023elite}, and InstantBooth~\cite{shi2024instantbooth} incorporate image encoders that extract and inject subject features into dense tokens via attention modules to achieve faster customization but rely on extensive multi-view training data, and may struggle to maintain identity consistency in out-of-distribution scenarios. 
Conversely, AnyDoor~\cite{chen2023anydoor} utilizes ID tokens and detailed maps to jointly represent subject features, facilitating more precise and zero-shot image generation. 
CustomNet~\cite{yuan2023customnet} integrates 3D novel view synthesis capabilities into the customization process, adeptly maintaining the object's identity. 
However, the aforementioned methods only support a single image as input. Our proposed approach accommodates multiple subject images, eliminating the necessity for individual prompts. By concatenating all subject images at the pixel level, we enable customized image generation that faithfully reflects the features of each subject. 
\section{Method}
\label{sec:method}

\begin{figure*}
  \centering
  \includegraphics[width=\textwidth]{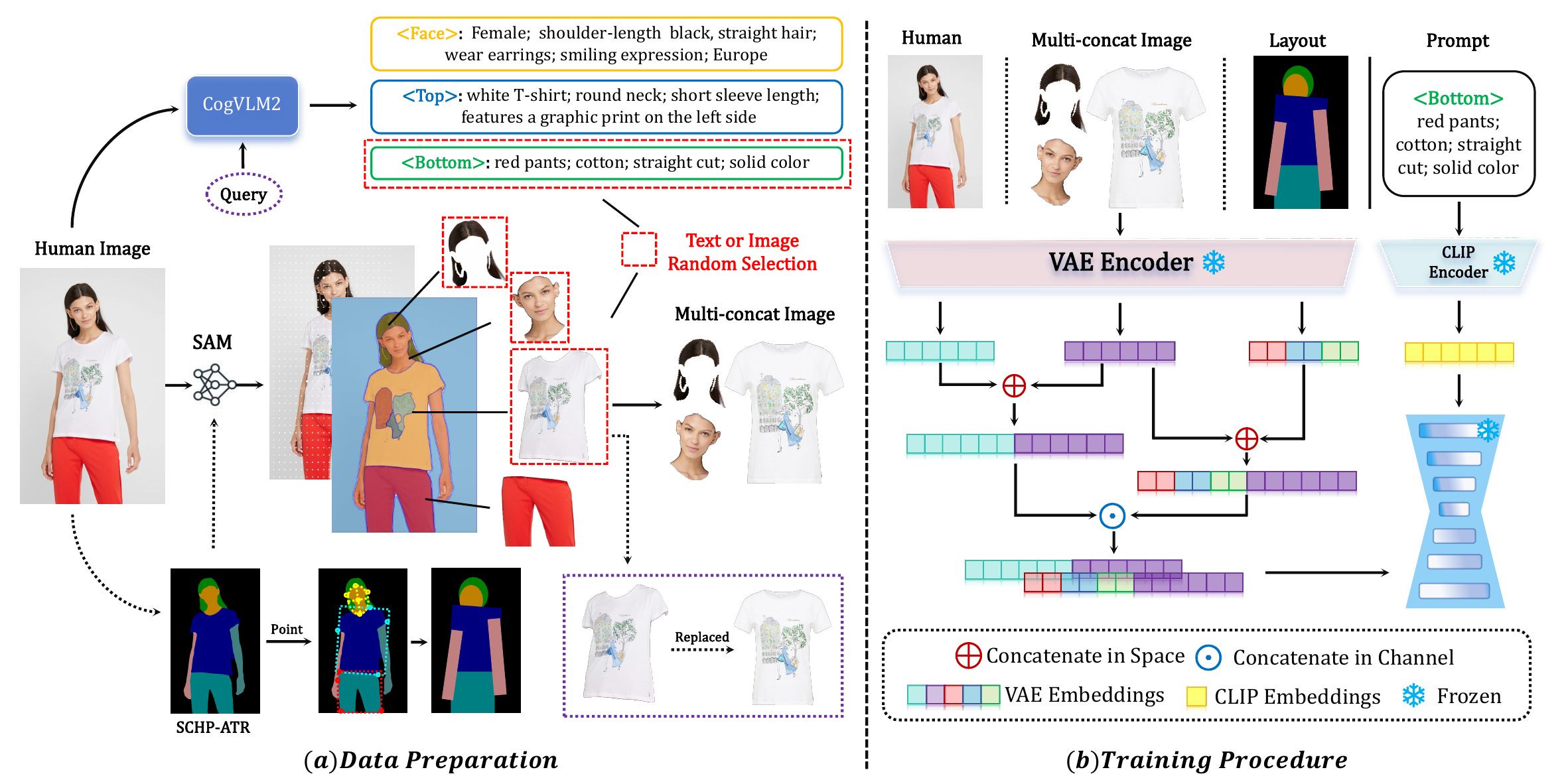}
  \vspace{-2mm}
  \caption{Overview of ComposeAnyone.
  $(a)$Data Preparation. We leverage CogVLM2, SAM, and SCHP to enrich the image-based try-on datasets with fine-grained textual descriptions, hand-drawn layouts, and human component sets. 
  $(b)$Training Procedure. We begin by employing the VAE encoder and the CLIP encoder to extract image and text embeddings, respectively, subsequently injecting the embeddings into the U-Net through concatenation across space and channel dimensions, yielding impressive results without the necessity of additional feature networks.
  }
  \vspace{-2mm}
  \label{fig:Training}
\end{figure*}

\subsection{Preliminaries}
\label{sec:pre}
\noindent \textbf{Stable Diffusion.} 
Stable Diffusion is a text-conditioned latent diffusion model~\cite{rombach2021highresolution}. For a VAE~~\cite{kingma2013auto} encoded image latent feature $z_0$, the forward diffusion process is performed by adding noise according to a noise scheduler $\alpha_t$~\cite{ho2020denoising}:
\begin{equation}
q(z_t|z_0) = \mathcal{N}(z_t; \sqrt{\alpha_t}z_0, (1 - \alpha_t)I).
\end{equation}

To reverse the diffusion process, a noise estimator $\epsilon_\theta(\cdot)$ parameterized by an UNet is learned to predict the forward added noise $\epsilon$ with the objective function,
\begin{equation}
\mathcal{L}_\text{dm} = \mathbb{E}_{(z_0,c)\sim D} \mathbb{E} _{\epsilon\sim\mathcal{N}(0,1),t}
\left[ ||\epsilon - \epsilon_\theta (z_t, t, c)||^2 \right],
\end{equation}
where $c$ is the text condition associated with image latent $z$, and $D$ is the training set. In the Stable Diffusion model, each block of the UNet is composed of cross-attention and self-attention layers. The cross-attention layer performs attention between the image feature query and the text condition embedding, and self-attention is performed within the image feature space. 


\subsection{Decoupled Multimodal Conditions}
\label{sec:condition}
\begin{figure}[t]
  \centering
  \includegraphics[width=0.95\linewidth]{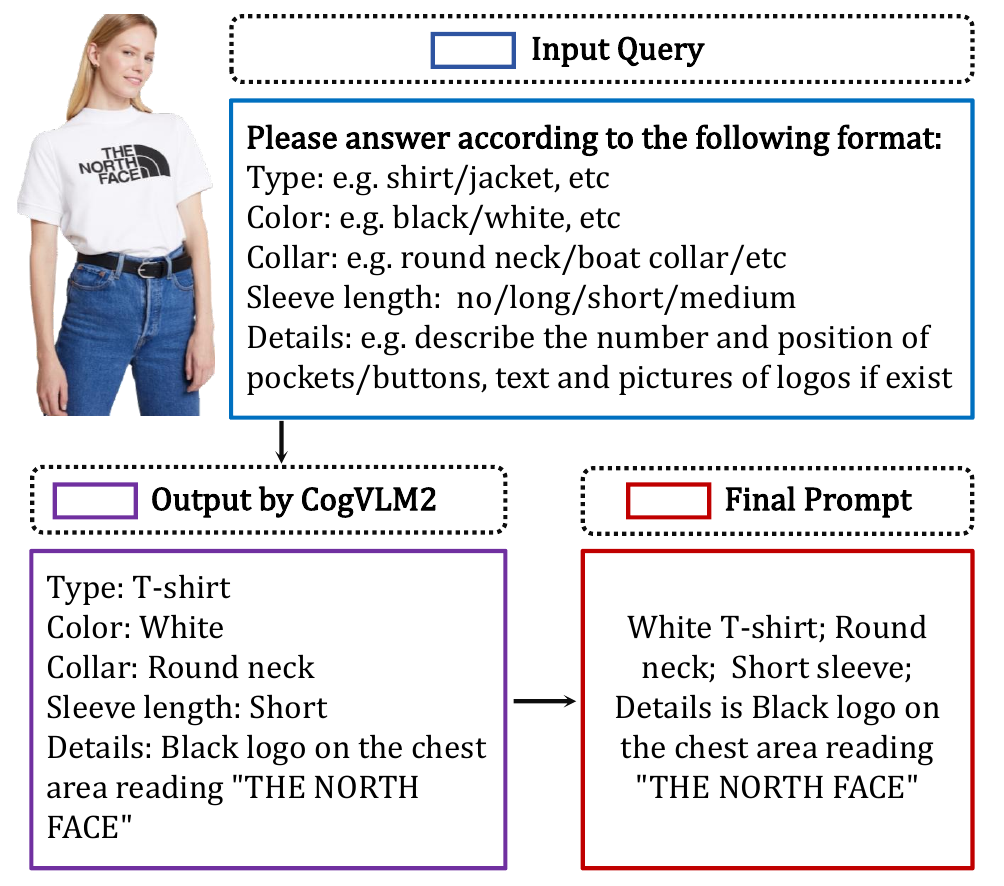}
   \vspace{-2mm}
   \caption{An example of fine-grained text description of $Top$. We use CogVLM2 to extract various attributes corresponding to components. Finally, these attributes are combined and transformed into a single descriptive sentence.}
   \label{fig:text}
   \vspace{-4mm}
\end{figure}

Existing methods~\cite{li2023gligen,ye2023ip-adapter,patel2024lambda,yuan2023customnet,wei2023elite} supporting joint image and text inputs rely on paired formats, which serve two primary purposes: (1) enhancing generative performance through mutual interaction via textual descriptions or specified categories, and (2) enabling text-driven editing or transformation of specific subject images. In contrast, we introduce a novel approach by decoupling image and text, aligning each modality independently.
Additionally, we incorporate hand-drawn layouts to provide further control over spatial arrangement. As illustrated in \Cref{fig:Training}(a), our method supports three distinct conditions for decoupled inputs:

\vspace{1mm}
\noindent \textbf{Text-Based Conditions.} 
Currently, vision language models (VLMs) exhibit remarkable proficiency in image annotation, offering high accuracy and exceptional clarity in descriptive detail.
We utilize CogVLM2~\cite{hong2024cogvlm2} to expand the fine-grained textual descriptions for each human image by applying component-level queries with standardized formatting to attributes such as face, top, bottom, and shoes.
These attributes are combined and transformed into a single descriptive sentence finally, as shown in \Cref{fig:text}. 

\vspace{1mm}
\noindent \textbf{Image-Based Conditions.} 
We extract distinct components of the human image, including hair, face, top, bottom, and shoes, to construct the reference image set. For human components segmentation, we integrated SAM~\cite{kirillov2023SAM} and SCHP~\cite{li2020schp} in a cross-validating way. We first calculate the SSIM~\cite{wang2004ssim} to assess the similarity between the masks of the components extracted by both methods. If the similarity exceeds 0.75, both extractions are deemed successful, with the smaller mask prioritized. Otherwise, indicating failure, the data is flagged for manual review or removed for cleaning.
We subsequently apply data augmentation, including rotation, flipping, and scaling, to the extracted components.

\vspace{1mm}
\noindent \textbf{Hand-Draw Layouts.} 
Based on SCHP~\cite{li2020schp}, we calculated contour coordinates for color blocks and fitted them into basic shapes such as ellipses and rectangles to build a coarse-grained hand-draw layout set.
Mathematically, 
the basic shapes are classified into ellipses and rectangles, a distinction that enhances user-friendliness and accessibility.
For the bounding ellipse, the fitting is performed using the least squares method. The center \( (x_c, y_c) \), semi-major axis \( a \), and semi-minor axis \( b \) are calculated as:
\begin{equation}
x_c = \frac{x_{\min} + x_{\max}}{2}, \quad y_c = \frac{y_{\min} + y_{\max}}{2} ,
\end{equation}
\begin{equation}
a = \frac{x_{\max} - x_{\min}}{2}, \quad b = \frac{y_{\max} - y_{\min}}{2} ,
\end{equation}
where \( (x_{\min}, y_{\min}) \) and \( (x_{\max}, y_{\max}) \) are the coordinates of the minimum bounding box surrounding the region.

For the rotated rectangle, the center \( (x_r, y_r) \), width \( w \), height \( h \), and rotation angle \( \theta_r \) are calculated. The coordinates of the rectangle's corners \( (x_i, y_i) \) are given by:
\begin{equation}
x_i = x_r + \cos(\theta_r + \alpha_i) \cdot d_i ,
\end{equation}
\begin{equation}
y_i = y_r + \sin(\theta_r + \alpha_i) \cdot d_i ,
\end{equation}
where \( \alpha_i \) is the angle from the center to each corner and \( d_i \) is the distance from the center to each corner.

\vspace{1mm}
Through the above strategies, we have completed the preparation for the three modalities, text, image, and layout, which constitute the foundation of our ComposeHuman dataset.
To achieve decoupling, we introduce a stochastic "$Text$ $or$ $Image$" input modality. We randomly select several reference images from the corresponding human component set and subsequently check the relevant labels to extract and discard the associated textual descriptions. This procedure establishes our comprehensive input condition.

\subsection{Controllable Layout-to-Human Generation}


Although numerous layout-to-image methods~\cite{densediffusion,bar2023multidiffusion,wang2024instancediffusion,zhou2024migc} exhibit remarkable performance, they predominantly rely on text as the sole auxiliary condition. Furthermore, most subject-driven approaches~\cite{ye2023ip-adapter,wei2023elite,yuan2023customnet,chen2023anydoor} are limited to single-image references. In contrast, our method facilitates the generation of multimodal, controllable humans with multiple image references, as shown in \Cref{fig:Training}.
Leveraging decoupled multimodal conditions, we facilitate the generation of humans in text-only, image-only, and mixed modalities by seamlessly concatenating along the spatial and channel dimensions.
The overall inputs consist of layout \( L_i = \{ l_1, l_2, \dots, l_n \} \), prompt \( T_i = \{ t_1, t_2, \dots, t_n \}\), reference Image set \( R_i = \{ r_1, r_2, \dots, r_n \}\) and Human Image \( H_i = \{ h_1, h_2, \dots, h_n \}\).
For inputs comprising multiple reference images, we apply a non-overlapping concatenation at the pixel level.
Then the images are passed through the VAE encoder~\cite{kingma2013auto}, mapping the images to a latent space:
\begin{equation}
z_{\text{human}} = E(h_i), \quad z_{\text{ref}} = E(r_i), \quad z_{\text{layout}} = E(l_i).
\end{equation}

The prompt \( T_i \) is translated by using a CLIP encoder~\cite{radford2021clip} to produce the text embedding:
\begin{equation}
\mathbf{t}_i = CLIP(t_i).
\end{equation}

The key step involves concatenating latent vectors along a chosen(-1) spatial dimension. Specifically, we concatenate the latents from the human image (\( z_{\text{human}} \)) and reference image (\( z_{\text{ref}} \)) to form the ground truth latent:
\begin{equation}
Z_{\text{gt}} = \text{Concat}(z_{\text{human}}, z_{\text{ref}}).
\end{equation}

Similarly, the source latent (from the layout image \( z_{\text{layout}} \)) is concatenated with the modified reference latent (where some parts may be set to zero based on a drop mask):
\begin{equation}
Z_{\text{src}} = \text{Concat}(z_{\text{layout}}, z_{\text{ref}}^{\text{drop}}).
\end{equation}


\begin{figure*}
  \centering
  \includegraphics[width=0.97\textwidth]{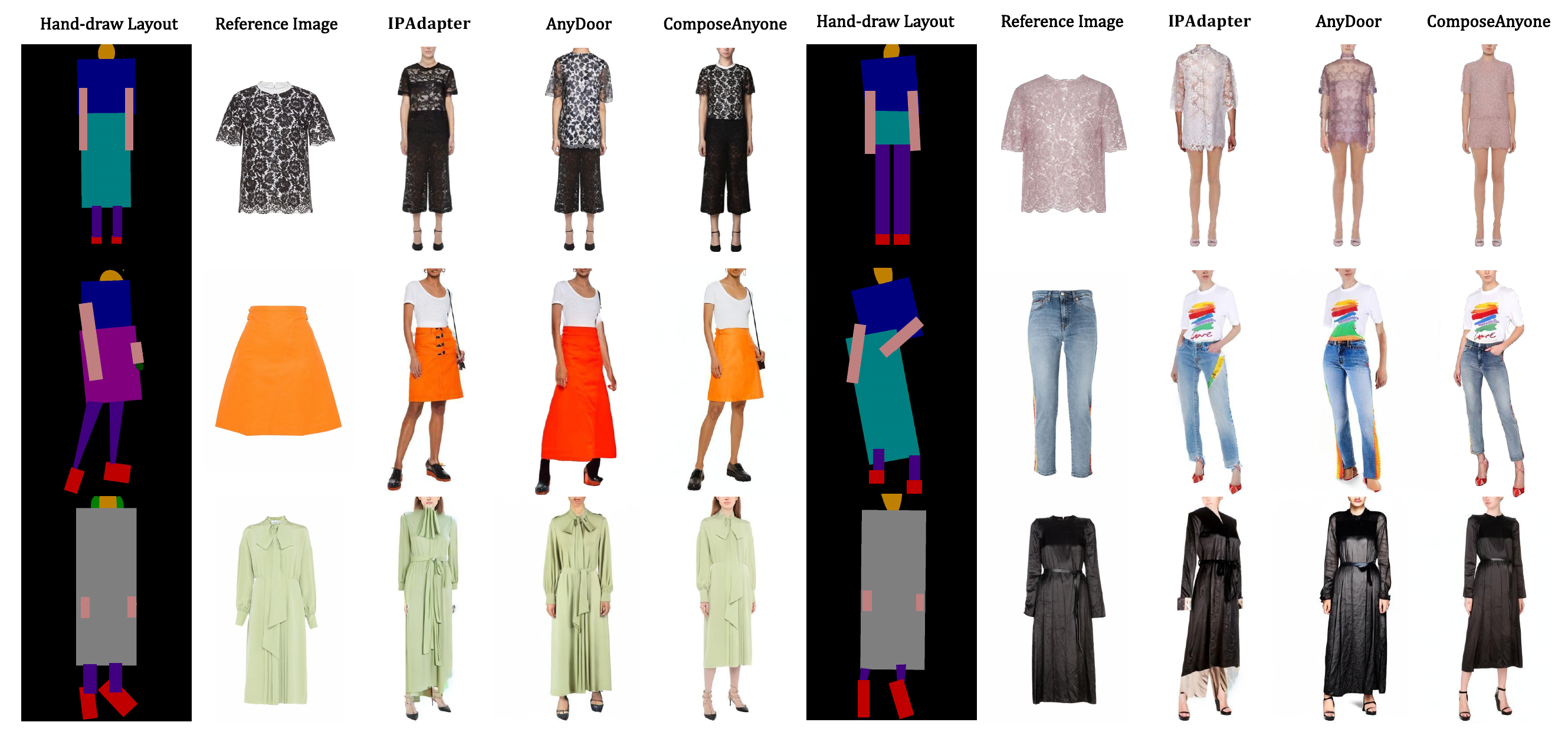}
    \vspace{-3mm}
  \caption{Qualitative comparison with subject-driven methods. ComposeAnyone demonstrates high fidelity in matching specific features of a given reference cloth image.}
  \label{fig:comparison1}
  \vspace{-3mm}
\end{figure*}

\begin{figure*}
  \centering
  \includegraphics[width=0.97\textwidth]{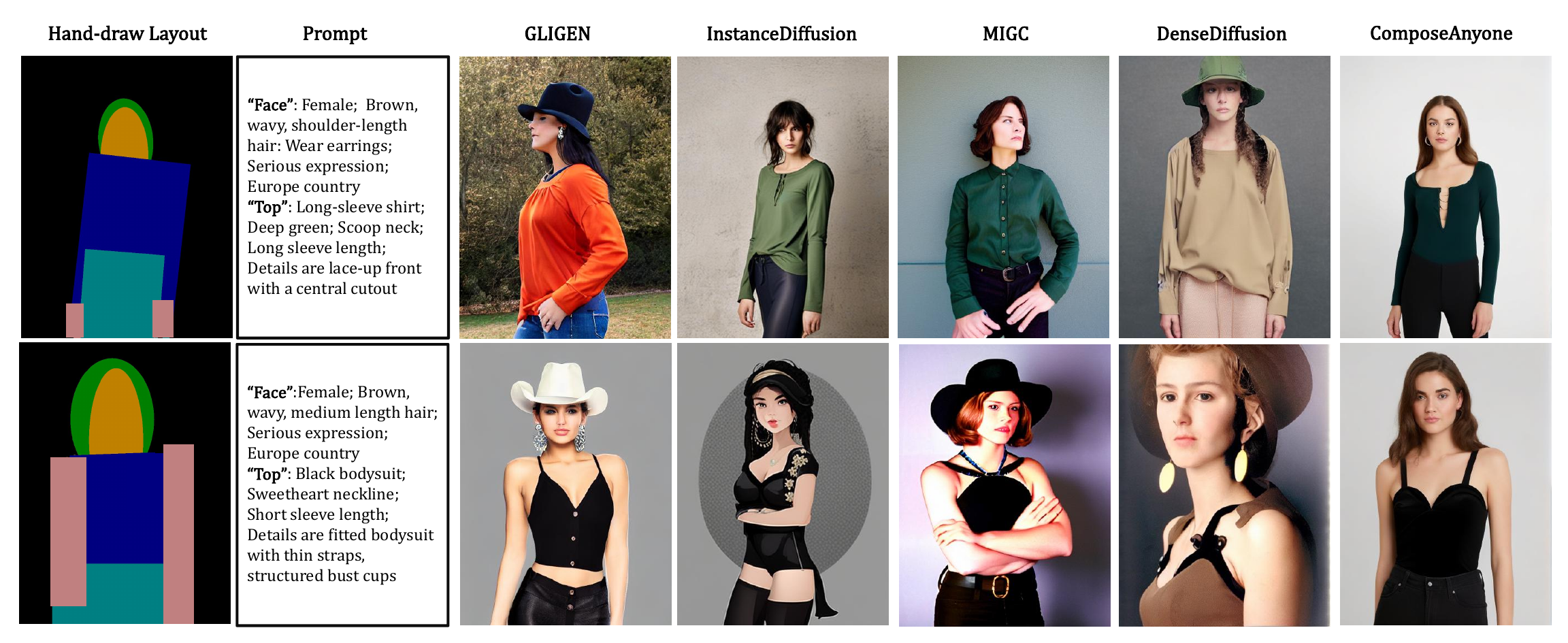}
    \vspace{-3mm}
  \caption{Qualitative comparison with layout-guided text-to-image methods. ComposeAnyone demonstrates a high level of congruity with both textual descriptions and spatial layout arrangements in its generative output.}
  \label{fig:comparison2}
  \vspace{-3mm}
\end{figure*}




The primary loss function is the Mean Squared Error (MSE) between the predicted noise \( \hat{\mathbf{\epsilon}} \) and the actual noise \( \mathbf{\epsilon} \):
\begin{equation}
L_{\text{MSE}} = \frac{1}{N} \sum_{i=1}^{N} \left( \hat{\mathbf{\epsilon}}_i - \mathbf{\epsilon}_i \right)^2 ,
\end{equation}
where \( N \) is the batch size.

Subsequently, we use signal-to-noise ratio (SNR) based loss weighting, enhancing the signal quality while mitigating the influence of noise. The loss is modified based on the SNR at each timestep:
\begin{equation}
L_{\text{total}} = L_{\text{MSE}} \cdot w_i ,
\end{equation}
where \( w_i \) is the weight based on the SNR for timestep \( t_i \), which is computed as:
\begin{equation}
w_i = \frac{1}{\text{SNR}(t_i)}  .
\end{equation}

Finally, the loss is backpropagated through the network:
\begin{equation}
\theta = \theta - \eta \nabla_\theta L_{\text{total}},
\end{equation}
where \( \theta \) represents the model parameters, \( \eta \) is the learning rate, and \( \nabla_\theta L_{\text{total}} \) is the gradient of the total loss concerning the parameters.

\begin{figure}[t]
  \centering
   \includegraphics[width=0.95\linewidth]{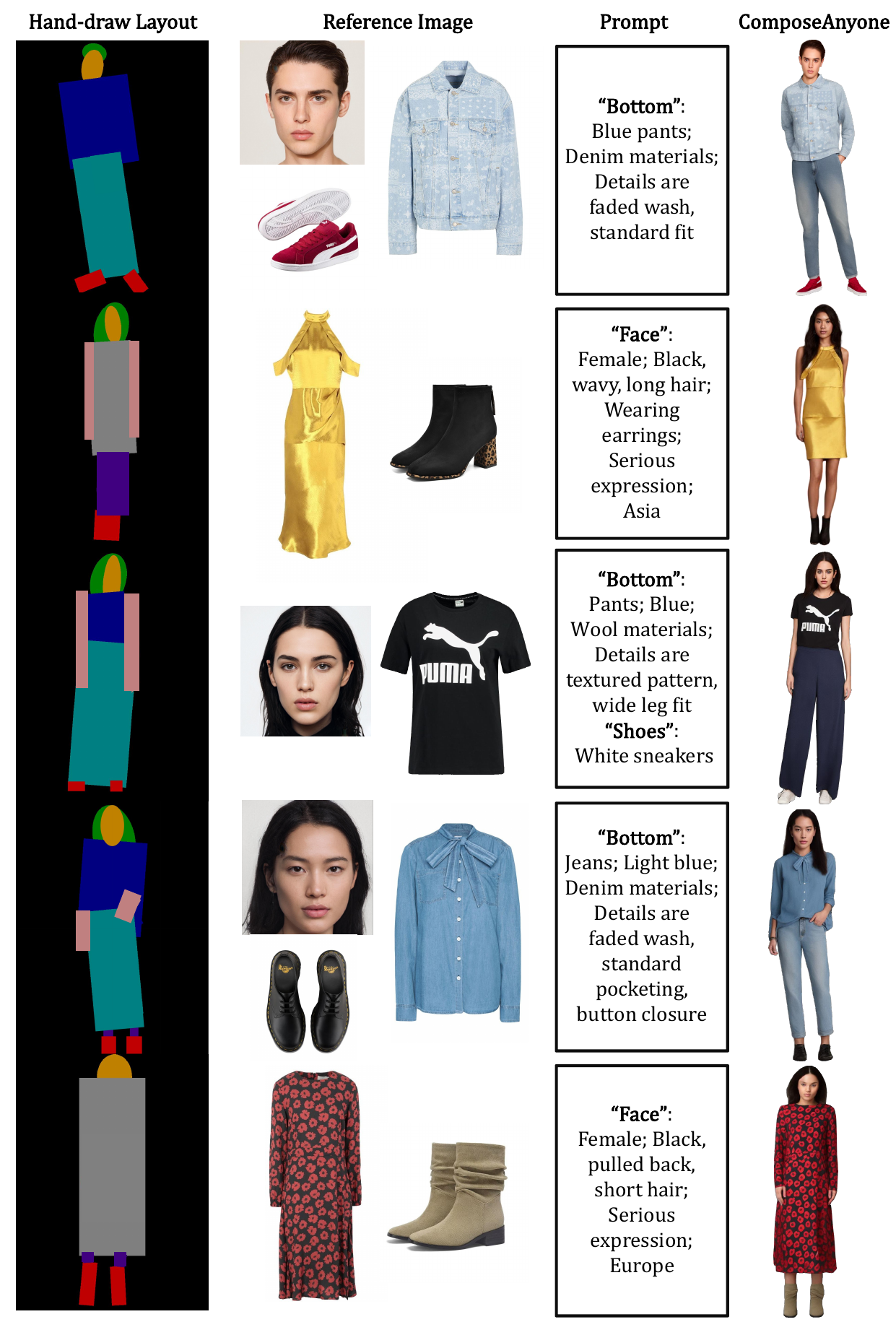}
   \vspace{-1mm}
\caption{Visual results of ComposeAnyone, highlighting its ability to process diverse modalities and generate high-quality human images that align with each input.}
   \label{fig:visual}
   \vspace{-4mm}
\end{figure}

\begin{figure}[t]
  \centering
   \includegraphics[width=\linewidth]{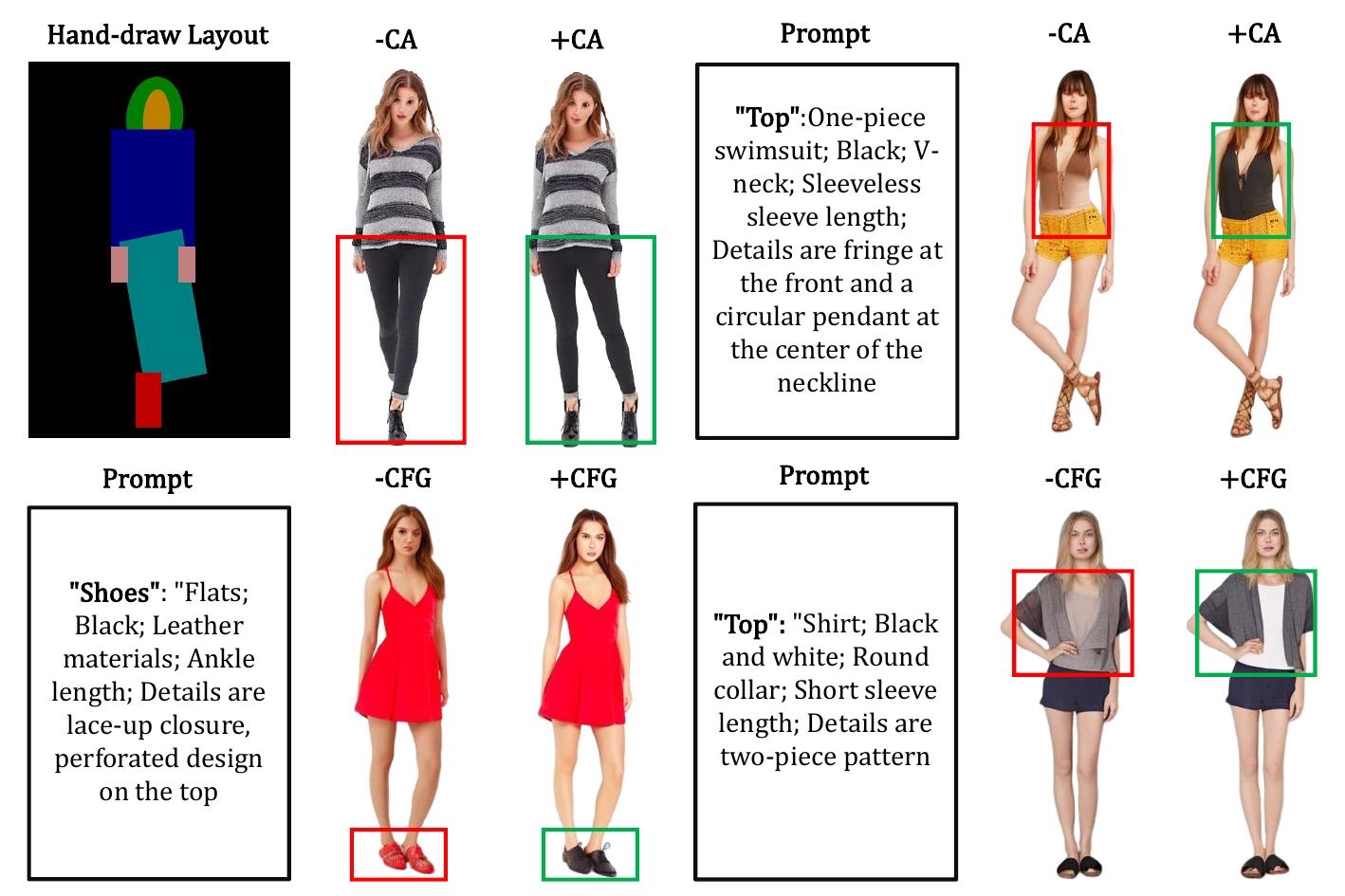}
    \vspace{-1mm}
   \caption{Ablation results of ComposeAnyone. The red boxes indicate incorrect generations, while the green boxes denote correct generations.}
   \label{fig:ablation}
   \vspace{-4mm}
\end{figure}

\begin{table*}[t]
    \centering
    \resizebox{\textwidth}{!}{
    \begin{tabular}{l|c|c|cccc|c}
        \toprule
        \multirow{2}{*}{Methods} & \multicolumn{1}{c|}{VLM Rate(\%)} & \multicolumn{1}{c|}{Spatial Accuracy(\%)} & \multicolumn{4}{c|}{Image Quality} & \multicolumn{1}{c}{Semantic Alignment}\\
        \cline{2-8}
         & CogVL $\uparrow$ & Average $\uparrow$  & SSIM\ $\uparrow$  & FID $\downarrow$ & KID $\downarrow$ & LPIPS $\downarrow$  & CLIP $\uparrow$ \\
        \midrule
        Real images & 83.95 & 49.05 & - & - & - & - & 28.6709   \\
        \hline
        GLIGEN~\cite{li2023gligen}  & 64.23 & \underline{41.01} & 0.3403 & \underline{87.0985} & \underline{70.6761} & 0.5862 & 27.7150   \\
        DenseDiffusion~\cite{densediffusion} & 67.37 & 32.46 & \underline{0.4177} & 97.9627 & 79.4878 & \underline{0.5605} & \underline{27.8251} \\
        MultiDiffusion~\cite{bar2023multidiffusion} & 64.52 & 25.87 & 0.2256 & 167.8483 & 141.3062 & 0.7296 & 21.1028   \\
        InstanceDiffusion~\cite{wang2024instancediffusion} & \underline{68.24} & 38.02 & 0.2840 & 113.6980 & 85.2871 & 0.6122 & 27.6839   \\
        MIGC~\cite{zhou2024migc}  & 67.07 & 38.97 & 0.3223 & 106.3653 & 93.3620 & 0.6048 & 26.1136  \\
        \hline
         ComposeAnyone (Ours) & \textbf{79.23} & \textbf{47.60} & \textbf{0.7381} & \textbf{18.3346} &  \textbf{9.9355} & \textbf{0.1553} & \textbf{28.2808}  \\

        \bottomrule
    \end{tabular}
    }
    \vspace{-2mm}
    \caption{Quantitative comparison with other layout-guided methods. We compare the metrics on the VITON-HD~\cite{choi2021vitonhd} datasets. The best and second-best results are demonstrated in \textbf{bold} and \underline{underlined}, respectively.}
    \label{tab:vitonhd_compare}
\end{table*}

Additionally, we use gradient clipping to avoid exploding gradients:
\begin{equation}
\nabla_\theta L_{\text{total}} = \text{Clip}(\nabla_\theta L_{\text{total}}, max\_grad\_norm)
\end{equation}
where $max\_grad\_norm$ is the threshold for gradient clipping.

During the inference step, we incorporate cross-attention control derived from hand-drawn layouts, thereby augmenting the quality and precision of text description generation. 
Let \( C = \{C_1, C_2, \dots, C_N\} \) represent the \( N \) color blocks in the hand-drawn layout $L$, where each color block \( C_i \) corresponds to a distinct key (e.g., face, top, bottom, shoes).
\( M_i \) is the binary mask for each color block, defined as:
\begin{equation}
M_i(x, y) = \begin{cases} 
1 & \text{if } (x, y) \in C_i \\
0 & \text{otherwise}
\end{cases}.
\end{equation}

The cross-attention map for each textual description is computed as follows:
\begin{equation}
A_{(i)} = \text{Attention}(\mathbf{Q}_i, \mathbf{K}_i, \mathbf{V}_i),
\end{equation}
where \( \mathbf{Q}_i \) is the query vector derived from the input image features. \( \mathbf{K}_i \) and \( \mathbf{V}_i \) are the key and value vectors corresponding to the textual description.

We modulate the cross-attention map \( A_{\text{cross}}^{(i)} \) using the binary mask \( M_i \) from the color block \( C_i \). This modulation adjusts the attention distribution by enhancing or suppressing attention in specific regions:
\begin{equation}
A_{(i)}^* = A_{(i)} \odot M_i,
\end{equation}
where \( \odot \) denotes element-wise multiplication, which increases or decreases attention in regions specified by the white areas of the mask.

The modulated cross-attention map $ A_{(i)}^* $ is injected into the model, guiding the attention mechanism and improving the generated textual descriptions. The model's final output is influenced by these modulated attention maps.

\begin{table*}[t]
    \centering
    \resizebox{\textwidth}{!}{
    \begin{tabular}{l|c|c|cccc|c}
        \toprule
       \multirow{2}{*}{Methods} & \multicolumn{1}{c|}{VLM Rate(\%)} & \multicolumn{1}{c|}{Spatial Accuracy(\%)} & \multicolumn{4}{c|}{Image Quality} & \multicolumn{1}{c}{Semantic Alignment}\\
        \cline{2-8}
         & CogVL $\uparrow$ & Average $\uparrow$  & SSIM\ $\uparrow$  & FID $\downarrow$ & KID $\downarrow$ & LPIPS $\downarrow$  & CLIP  $\uparrow$\\
        \midrule
        Real images & 82.75 & 62.80 & - & - & - & - & 25.9114 \\
        \hline
        IP-Adapter~\cite{ye2023ip-adapter}  & \underline{77.78} & 16.72 & 0.4937 & 57.5815 & 36.1861 & 0.5521 & 24.8649\\
        $\lambda$-ECLIPSE~\cite{patel2024lambda} & 71.30 & 38.93 & 0.5999 & 33.8015 & 17.7533 & 0.4097 & \underline{25.5986}\\
        ELITE~\cite{wei2023elite} & 72.56 & 26.15 & 0.4138 & 119.7391 & 90.6251 & 0.5784 & 22.8187\\
        CustomNet~\cite{yuan2023customnet} & 68.11 & 38.32 & 0.6620 & 71.6908 & 45.3029 & 0.3435 & 24.7704\\
        AnyDoor~\cite{chen2023anydoor}  & 71.91 & \underline{42.16} & \underline{0.7347} & \underline{32.5295} & \underline{17.4754} & \underline{0.2165} &  24.9714\\
        \hline
         ComposeAnyone (Ours) & \textbf{79.18} & \textbf{61.77} & \textbf{0.8095} & \textbf{11.6239} & \textbf{3.5040} & \textbf{0.1461} &  \textbf{25.8175}\\

        \bottomrule
    \end{tabular}
    }
    \vspace{-2mm}
    \caption{Quantitative comparison with other subject-driven methods. We compare the metrics on the DressCode~\cite{morelli2022dresscode} datasets. The best and second-best results are demonstrated in \textbf{bold} and \underline{underlined}, respectively.}
    \label{tab:dresscode_compare}
    \vspace{-2mm}
\end{table*}

\begin{table*}[t]
    \centering
    \resizebox{0.95\textwidth}{!}{
    \begin{tabular}{l|c|c|cccc|cccc}
        \toprule
        \multirow{2}{*}{Version} & \multicolumn{1}{c|}{VLM Rate(\%)} & \multicolumn{1}{c|}{Spatial Accuracy(\%)} & \multicolumn{4}{c|}{Image Quality} & \multicolumn{4}{c}{Semantic Alignment $\uparrow$}\\
        \cline{2-11}
        & CogVL $\uparrow$ & Average $\uparrow$  & SSIM\ $\uparrow$  & FID $\downarrow$ & KID $\downarrow$ & LPIPS $\downarrow$  & Face & Top & Bottom & Shoes\\
        \midrule
        Real images & 81.04 & 69.22 & - & - & - & - & 22.9165 & 29.5885 & 27.8829 & 22.8722\\
        \hline
        $-CA  -CFG$ & 76.99 & 68.49 & 0.8539 & 17.0397 & 3.2297 & 0.1003 & 22.6219 & 29.6539 & 27.7048 &  22.7530 \\
        $+CA  -CFG$ & 78.04 & 69.62 & 0.8220 & 19.9730 & 6.2259 & 0.1209 & 23.2320 & 29.3456 & 27.8596 & 22.8890 \\
        $-CA  +CFG$ & 80.35 & 67.76 & 0.8690 & 14.8462 & 1.8699 & 0.0735 & 22.6828 & 29.6166 & 27.8558 & 22.8827 \\
        $+CA  +CFG$ & 80.61 & 69.58 & 0.8171 & 20.2401 & 6.3680 & 0.1242 & 23.2407 & 29.3409 & 27.8710 & 22.8829 \\
        \bottomrule
    \end{tabular}
    
    }
    \vspace{-2mm}
    \caption{Ablation results of ComposeAnyone. We compare the metrics on the DeepFashion~\cite{DeepFashion2} dataset.}
    \label{tab:ablation}
    \vspace{-2mm}
\end{table*}


\section{Experiments}
\label{sec:experiment}
\subsection{Datasets}
\label{sec:datasets}
Our training dataset was sampled from three publicly available image-based virtual try-on datasets—VITON-HD~\cite{choi2021vitonhd}, DressCode~\cite{morelli2022dresscode}, and DeepFashion~\cite{DeepFashion2}—comprising 9,027, 48,392, and 5,592 front-view image pairs, respectively, forming the ComposeHuman dataset following strategies mentioned in \Cref{sec:condition}.
For robust efficacy evaluation, we conducted comparative experiments using the VITON-HD~\cite{choi2021vitonhd}, DressCode~\cite{morelli2022dresscode}, and DeepFashion~\cite{DeepFashion2} test sets, containing 2,032, 2,000, and 1,000 samples, respectively.

\subsection{Implementation Details}
\label{sec:implementation}
We initialized our model's backbone with the weights from InstructPix2Pix~\cite{brooks2022instructpix2pix}, which is based on Stable Diffusion~\cite{rombach2021highresolution}. We trained the model at a resolution of 384×512 with a batch size of 16 for 52,000 steps. To further improve visual quality, we fine-tuned the model at a higher resolution of 768×1024 for an additional 46,000 steps. During training, we used the AdamW optimizer with a learning rate of 1e-4. All experiments were conducted on 4 NVIDIA A100 GPUs.

\subsection{Evaluation Metrics} 
\label{sec:metrics}

For validating layout-guided multimodal input generation, the model must generate instances with precise features at the specified spatial positions.

\noindent \textbf{VLM Rate.} Since CLIP may not capture intricate details effectively, we use CogVLM2~\cite{hong2024cogvlm2} to query and assess the alignment of instance features, facilitating more sophisticated, detailed, and varied evaluation.

\noindent \textbf{Spatial Accuracy.} We first use GroundingDINO~\cite{liu2023grounding} to detect human component positions in generated images and then compute the Jaccard~\cite{jaccard1901etude}, Dice~\cite{dice1945measures}, and SSIM~\cite{wang2004ssim} metrics between the detected positions and the hand-drawn layout, weighted with 0.25, 0.25, and 0.50, respectively.

\noindent \textbf{Image Quality.} To comprehensively evaluate image quality, we employ FID~\cite{Seitzer2020FID} and KID~\cite{bińkowski2021kid} to assess fidelity by calculating feature similarity between image sets, and SSIM~\cite{wang2004ssim} and LPIPS~\cite{zhang2018LPIPS} to evaluate structural similarity between individual images.

\noindent \textbf{Semantic Alignment.} We apply CLIP Score~\cite{taited2023CLIPScore} to evaluate the relevance between component-level text and the generated images across four distinct attributes: face, top, bottom, and shoes. The final score is the average of these four individual scores.

\subsection{Qualitative Comparison}

To demonstrate the efficacy of our approach, we conducted qualitative analyses with subject-driven~\cite{chen2023anydoor,ye2023ip-adapter} and layout-guided~\cite{li2023gligen,wang2024instancediffusion,zhou2024migc,densediffusion} methods, as shown in \Cref{fig:comparison1} and \Cref{fig:comparison2}. Although GLIGEN~\cite{li2023gligen}, InstanceDiffusion~\cite{wang2024instancediffusion}, and MIGC~\cite{zhou2024migc} approximate spatial layouts, they struggle to accurately control human pose, with InstanceDiffusion~\cite{wang2024instancediffusion} frequently producing unrealistic figures and DenseDiffusion~\cite{densediffusion} often neglecting key semantic details. ComposeAnyone achieves better alignment with both the layout specifications and textual inputs. Additionally, when given a garment reference image, ComposeAnyone consistently captures intricate visual details, outperforming AnyDoor~\cite{chen2023anydoor} and IP-Adapter~\cite{ye2023ip-adapter}. \Cref{fig:visual} further demonstrates ComposeAnyone's ability to generate lifelike human images that seamlessly align with multimodal inputs.


\subsection{Quantitative Comparison}

\noindent \textbf{Layout-Guided Text-to-Human Generation.} 
We conducted quantitative comparisons with state-of-the-art layout-guided text-to-image methods~\cite{li2023gligen,densediffusion,bar2023multidiffusion,wang2024instancediffusion,zhou2024migc} on the VITON-HD~\cite{choi2021vitonhd} dataset. As shown in \Cref{tab:vitonhd_compare}, our proposed ComposeAnyone outperforms the baseline models across all metrics listed in \Cref{sec:metrics}, generating spatially coherent human images that align with the hand-drawn layout and consistently match the textual descriptions. This demonstrates its robustness and versatility.

\noindent \textbf{Subject-Driven Human Generation.} We conducted quantitative comparisons with advanced subject-driven image generation methods~\cite{ye2023ip-adapter,patel2024lambda,wei2023elite,yuan2023customnet,chen2023anydoor} on the DressCode~\cite{morelli2022dresscode} dataset. As shown in \Cref{tab:dresscode_compare}, our method achieves superior performance across all metrics. This highlights ComposeAnyone's ability to robustly retain subject features while ensuring precise positioning of components in the target image.

\subsection{Ablation Study}
We conducted ablation studies on the DeepFashion\cite{DeepFashion2} dataset, focusing on classifier-free guidance (CFG) and cross-attention modulation (CA). As illustrated in \Cref{fig:ablation} and \Cref{tab:ablation}, the CA markedly enhances the model's ability to control textual input and refine facial details while also facilitating more precise spatial alignment. Moreover, the CFG bolsters the model's overall robustness. 
\section{Conclusion}
In this work, we introduce ComposeAnyone, a novel controllable Layout-to-Human generation method that combines hand-drawn geometric layouts with text and image references to generate high-quality, realistic human images across various modalities. Through the construction of the ComposeHuman dataset and multimodal decoupling, we have provided a more flexible and detailed approach to spatial organization in human image generation. Extensive experiments on layout-guided and subject-driven tasks demonstrate the effectiveness and robustness of our method, outperforming existing approaches in terms of alignment, fidelity, and adaptability.

\vspace{2mm}
\noindent \textbf{Limitation.}
While our method yields high-quality human images, it is not without inherent limitations. The training data used in our approach is derived from semantic segmentation models and vision-language models (VLMs), both of which may introduce certain inaccuracies. Additionally, biases present in pre-trained models can affect the robustness and reliability of the outputs, potentially compromising their adaptability to diverse user needs and contexts.
\label{sec:conclusion}

{
    \small
    \bibliographystyle{ieeenat_fullname}
    \bibliography{main}
}


\end{document}